\documentclass[journal]{IEEEtran}
\ifCLASSINFOpdf
   \usepackage[pdftex]{graphicx}
\else
\fi
\usepackage[cmex10]{amsmath}
\usepackage{multirow}
\usepackage{subcaption}
\usepackage{float}
\usepackage[colorlinks=true,breaklinks=true,bookmarks=true,urlcolor=blue,citecolor=blue,linkcolor=blue,bookmarksopen=false,draft=false]{hyperref}

\usepackage[sort&compress,numbers]{natbib}

\begin{document}
%
\title{DAWN: Vehicle Detection in Adverse Weather Nature }

%

\author{Mourad A. Kenk, Mahmoud Hassaballah
	\thanks{Manuscript received Feb. 27, 2020; Accepted: XXX, Published: XXXX. This paper was recommended by Associate Editor XYZ. (Corresponding author: Mourad A. Kenk)}
	\thanks{Mourad A. Kenk is with Department of Mathematics, Faculty of Science, South Valley University, Qena 83523, Egypt (e-mail: mourad.ahmed@sci.svu.edu.eg)}
	\thanks{Mahmoud Hassaballah is with the Department of Computer Science, Faculty of Computers and Information, South Valley University, Qena 83523, Egypt (e-mail: m.hassaballah@svu.edu.eg)}
}

\markboth{}%
{Mourad A. Kenk and M. Hassaballah \MakeLowercase{\textit{et al.}}: Bare Demo of IEEEtran.cls for Journals}

\maketitle

\begin{abstract}
Recently, self-driving vehicles have been introduced with several automated features including lane-keep assistance, queuing assistance in traffic-jam, parking assistance and crash avoidance. These self-driving vehicles and intelligent visual traffic surveillance systems mainly depend on cameras and sensors fusion systems. Adverse weather conditions such as heavy fog, rain, snow, and sandstorms are considered dangerous restrictions of the functionality of cameras impacting seriously the performance of adopted computer vision algorithms for scene understanding (i.e., vehicle detection, tracking, and recognition in traffic scenes). For example, reflection coming from rain flow and ice over roads could cause massive detection errors which will affect the performance of intelligent visual traffic systems. Additionally, scene understanding and vehicle detection algorithms are mostly evaluated using datasets contain certain types of synthetic images plus a few real-world images. Thus, it is uncertain how these algorithms would perform on unclear images acquired “in the wild” and how the progress of these algorithms is standardized in the field. To this end, we present a new dataset (benchmark) consisting of real-world images collected under various adverse weather conditions called DAWN. This dataset emphasizes a diverse traffic environment (urban, highway and freeway) as well as a rich variety of traffic flow. The DAWN dataset comprises a collection of 1000 images from real-traffic environments, which are divided into four sets of weather conditions: fog, snow, rain and sandstorms. The dataset is annotated with object bounding boxes for autonomous driving and video surveillance scenarios. This data helps interpreting effects caused by the adverse weather conditions on the performance of vehicle detection systems.

\end{abstract}


%
\IEEEpeerreviewmaketitle

\section{Introduction}
\IEEEPARstart{T}{he} efficiency of vehicle detection is considered as a critical step in traffic monitoring or intelligent visual surveillance in general \cite{zhang2018vehicle,liu2018dynamic}. Recently, the evolution of sensors and GPU along with deep learning algorithms has concentrated research into autonomous or self-driving applications based on artificial intelligence and became a trend \cite{chen2019surrounding}. Autonomous vehicles must precisely detect traffic objects (e.g., cars, cyclists, traffic lights, etc.) in real-time to right control decisions and ensure the required safety \cite{wu2017squeezedet}. To detect such objects, diverse sensors such as cameras and light detection and ranging are commonly utilized in autonomous vehicles. Among these various types of sensors, the quality of camera's images is quite affected by adverse weather conditions such as heavy foggy, sleeting rain, snowstorms, dusty blast, and low light conditions. Consequently, the visibility is inefficient for detecting accurately the vehicles on the roads and yields traffic accidents. Clear visibility can be reach by developing efficient image enhancement methods to obtained good visual appearance or discriminative features. Thus, providing detection systems with clear images can improve the performance of vehicle detection and tracking in intelligent visual surveillance systems and autonomous vehicles applications \cite{Cho2018Robot,Kuang2017Score, dong2019efficient}. 

Recently, computer vision community introduced different vehicle detection approaches \cite{min2018new}. In particular, deep learning based traffic object detection using camera sensors has become more significant in autonomous vehicles because it achieves high detection accuracy, and consequently, it has become a substantial method in self-driving applications \cite{Hu2019SINet}. Two essential conditions should be satisfied by the detector: a real time detection is necessary for an active echo of vehicle's controllers, and the high detection accuracy of the traffic objects is mandatory which has not been investigated under adverse weather conditions before. 

Although these methods have achieved fast detection with high efficiency, they could not improve the detection accuracy \cite{hassaballah2020local, cai2016unified}. Lately, object detectors based on CNN models that integrate various strategies have been widely studied to take advantage of both types of deep learning categories and to compensate for their particular drawbacks. CFENet \cite{zhao2018comprehensive}, a one-stage detector, has used an extensive feature improvement strategy based on SSD to increase the detection accuracy. RefineDet \cite{zhang2018single}, a one-stage detector, improves the detection accuracy by using an anchor refinement strategy and an object detection module. RFBNet \cite{liu2018receptive}, has applied a receptive field block to improve the detection accuracy. However, using hard lighting conditions without the presence of adverse weather conditions and with an input image resolution of $512\times512$ or higher have been incapable to achieve a real time detection speed above 30 frames per second as reported in previous studies \cite{Hu2019SINet, cai2016unified, zhao2018comprehensive, zhang2018single}. Real time detection is a requirement for traffic monitoring and self-driving applications under adverse weather conditions. Though, real time detection speed is achieved in \cite{liu2018receptive}, it is hard to employ it in adverse weather conditions because of low detection accuracy. This denotes that the previous strategies are insufficient in terms of a trade-off between the accuracy and time of detection, which restricts usage in applications with adverse weather conditions. It can confuse in the judgment of an accurate vehicle detection and reduce the efficiency of vehicles detection under adverse weather conditions and lead to a traffic accident. In other words, it is extremely important to employ a vehicle detector with high detection accuracy and consider this factor along with the real time detection speed to reduce the false alarms of the detected bounding boxes and to allow space of time to improve the visibility in the traffic environment under adverse weather conditions and thus preventing traffic accidents.

The available vehicle datasets in literature still need to address more challenging adverse weather conditions datasets. Table \ref{datastable} summarizes the available vehicles datasets in literature, where the datasets are collected by traffic surviellance camera (TSC), On-raod vehicles camera (OVC), web servey (Web), or by drone camera. On the other hand, there is no generic datasets for the different adverse weather conditions such as the combination of nasty winter weather, sleeting rain, and dust storms. For instance, Sakaridis et al. \cite{sakaridis2018semantic} proposed a convolution neural network (CNN) based model to generate synthetic fog on real vehicle images to investigate defogging algorithms in the traffic environments. Hodges et al. \cite{hodges2019single} manipulated the dehazing model by a dehazing network to reform the full image and a discriminator network to fine tunning the enhancement weights parameters to increase the vehicle detection performance on a dataset of synthetic foggy/hazy images. Li et al. \cite{li2019single} presented a benchmark including both synthetic and real-world rainy images with some rain types to investigate deraining algorithms in traffic monitoring scene and vehicle detection. Uzun et al. \cite{uzun2019cycle} implemented cycle-spinning with generative adversarial networks (GAN) for raindrops removal in outdoor surveillance systems and investigated the object detection performance under Raindrop dataset \cite{qian2018attentive}. However, these methods are mainly evaluated on rendered synthetic fog/rain images and few real images assuming a specific fog/rain model. It is thus unclear how these algorithms would be proceeding on various adverse weather conditions and how the progress could be measured in the wild.

To solve the problem, a new benchmark dataset is introduced called DAWN consisting of real world images collected under various adverse weather conditions (e.g., fog, rain, snow, and sandstorms). The collected images provide a diverse traffic environment (e.g., urban, crossroads, motorway, etc.) with various vehicles categories that are annotated for intelligent visual surveillance, traffic monitoring and self-driving vehicles applications.

\section{Background}
In this section, we present performance analysis of the proposed methods under different adverse weather conditions adapted on detecting vehicles categories (e.g., car, bus, truck, motorcycle, bicycle) with the presence of human as a category (person) for pedestrian and cyclist in traffic environment scenes.

\subsection{Summary of the available vehicles image datasets in literature.}
In this part we provide an overview of the datasets used for evaluating vehicle detector models, as the detail of the proposed dataset for vehicle detection in adverse weather.
\begin{table}[!]
\centering
\setlength{\tabcolsep}{4pt}
\caption{Summary of available vehicles datasets in literature. Lighting variations as $L$, Occlusion as $O$, Crowded as $C$.}
\label{datastable}
\begin{tabular}{|c|c|c|c|c|c|c|c|c|}
\hline 
Dataset & Mode & No. Image & Video & Test & Train & $L$ & $O$ & $C$ \\ 
\hline 
UA-DETRAC \citep{UADETRAC2020}& TSC & • & 10 Hour & $\times$ & • & $\times$ & $\times$ & $\times$ \\ 
\hline 
TME Motorway \citep{TMEMotorwayDataset} & OVC & • & 28 Clip & $\times$ & • & $\times$ & • & $\times$ \\ 
\hline 
KITTI \citep{geiger2012we} & OVC & $\times$ & • & 7518 & 7481 & • & $\times$ & $\times$ \\ 
\hline 
Stanford car \citep{krause20133d}& web & 16,185 & • & 50-50 & 50-50 & • & • & • \\ 
\hline 
PASCAL VOC \citep{pascal_voc2012}& web & $\times$ & • & $\times$ & $\times$ & • & $\times$ & $\times$ \\ 
\hline 
Rain\&Snow \citep{bahnsen2018rain}& TSC & • & 22 Clip & $\times$ & • & $\times$ & • & • \\ 
\hline 
Cityscape \citep{Cordts2016Cityscapes}& OVC & 25,000 & • & $\times$ & $\times$ & $\times$ & $\times$ & $\times$ \\ 
\hline 
Mapillary \citep{neuhold2017mapillary}& OVC & 25,000 & • & $\times$ & $\times$ & $\times$ & $\times$ & $\times$ \\ 
\hline 
BDD100K \citep{yu2018bdd100k}& OVC & 100,000 & • & $\times$ & $\times$ & $\times$ & $\times$ & $\times$ \\ 
\hline 
ApolloScape \citep{huang2018apolloscape}& OVC & 143,906 & • & $\times$ & $\times$ & $\times$ & $\times$ & $\times$ \\ 
\hline 
Stanford Drone \citep{robicquet2016learning}& drone & $\times$ & • & $\times$ & • & • & • & • \\ 
\hline 
\end{tabular} 
\end{table}
\subsubsection{KITTI dataset \cite{geiger2012we}}
\begin{figure}[!]
\centering
		\includegraphics[width=9 cm,height=!]{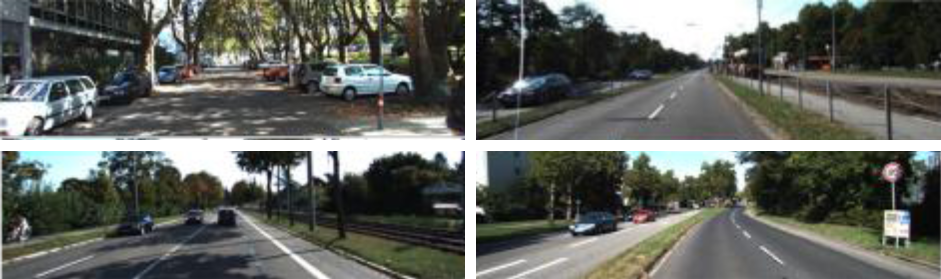}
		\caption{Sample images of the KITTI dataset.}
	\label{kitti_ex} 
\end{figure}

is the most widely used for on-road vehicle detection and self-driving researches. The KITTI dataset consists of 7,481 images for training and 7,518 images for testing and includes six classes: car, van, truck, tram, cyclist, and pedestrian. The input image size is 512$\times$512 and 17,607 total bounding box of GT. The KITTI dataset considers a traffic environment that covers freeways through rural zones and urban scenes with lighting variability in normal weather conditions at daylight only as shown in Figure. \ref{kitti_ex}.
\subsubsection{The MS-COCO dataset \cite{lin2014microsoft}}
\begin{figure}[!]
\centering
		\includegraphics[width=9 cm,height=!]{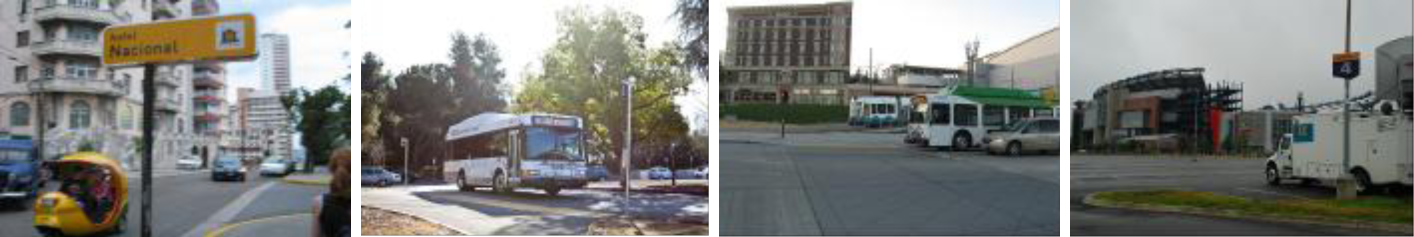}
		\caption{Sample images for vehicles in MS-COCO dataset.}
	\label{coco_ex} 
\end{figure}

is a more challenging scene understanding than KITTI dataset. It is often used by the current state-of-the-art deep learning models. MS-COCO includes a large-scale of complex scenes annotated for 80 classes where the traffic objects and environment scenes are addressing general settings for normal weather situations as shown in Figure. \ref{coco_ex}. 

\begin{figure}[!]
\centering%
\includegraphics[width=9cm,height=!]{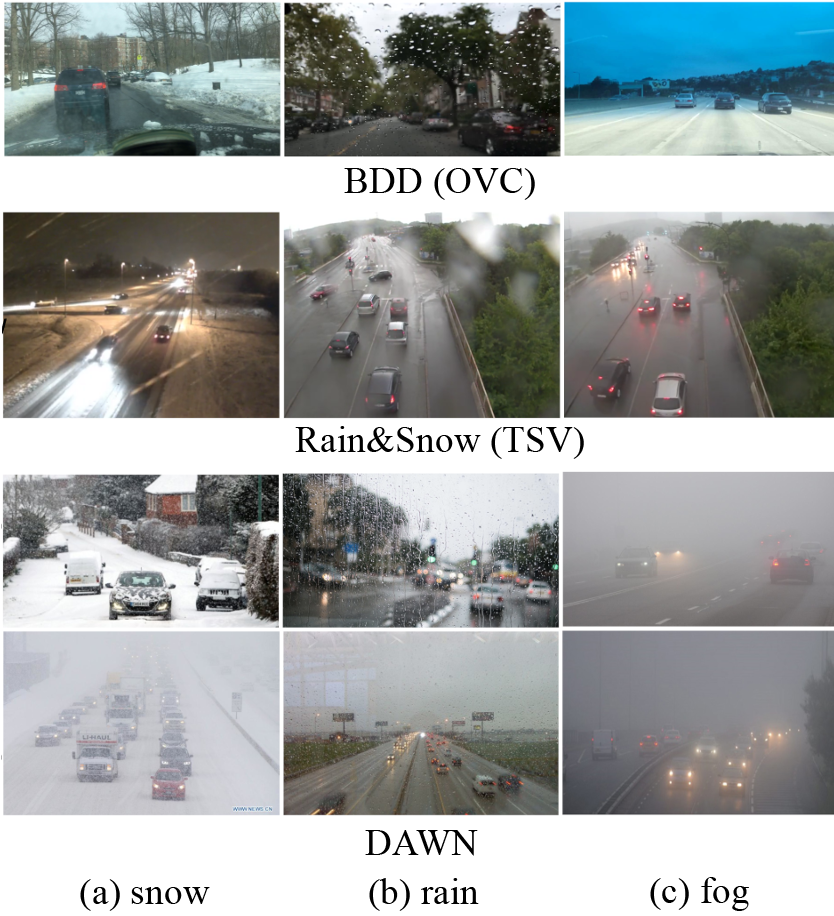}
\caption{Sample of traffic images for comparison with state of the art datasets (Rain\&Snow and BDD).}
\label{datasets_compr} 
\end{figure}

In these two datasets, the traffic scene is often addressing normal weather conditions. Moreover, we clarify that the top-performing methods for vehicle detection and visual scene understanding do not completely apprehend the difficulty and variability of poor real-world weather conditions. For more details on scene understanding datasets, we refer the readers to \cite{minaee2020image}. The disparity of traffic images in DAWN dataset and the state of the art datasets (Rain \& Snow and BDD) is compared and shown in Figure \ref{datasets_compr}. DAWN dataset include extrem level of weather condition and variation of traffic environments. In addition, the dataset is annotated with object bounding boxes for autonomous driving and video surveillance scenarios. 

\subsubsection{DAWN dataset}
To the best of our knowledge, few datasets address the problem of adverse weather conditions by certain types of synthetic weather in images plus a few real-world images. For instance, Sakaridis et al. \cite{sakaridis2018semantic} proposed two datasets; synthetic foggy cityscapes and foggy driving datasets to investigate vehicle detection and defogging algorithms in traffic environments with 8 classes. Li et al. \cite{li2019single} introduced a benchmark to evaluate deraining algorithms in the traffic scene consisting of rain in driving and surveillance datasets. This dataset consists of synthetic and real-rainy environment of 2,495 and 2,048 images, respectively.  
\begin{figure*}[!]
\centering
		\includegraphics[width=18 cm,height=!]{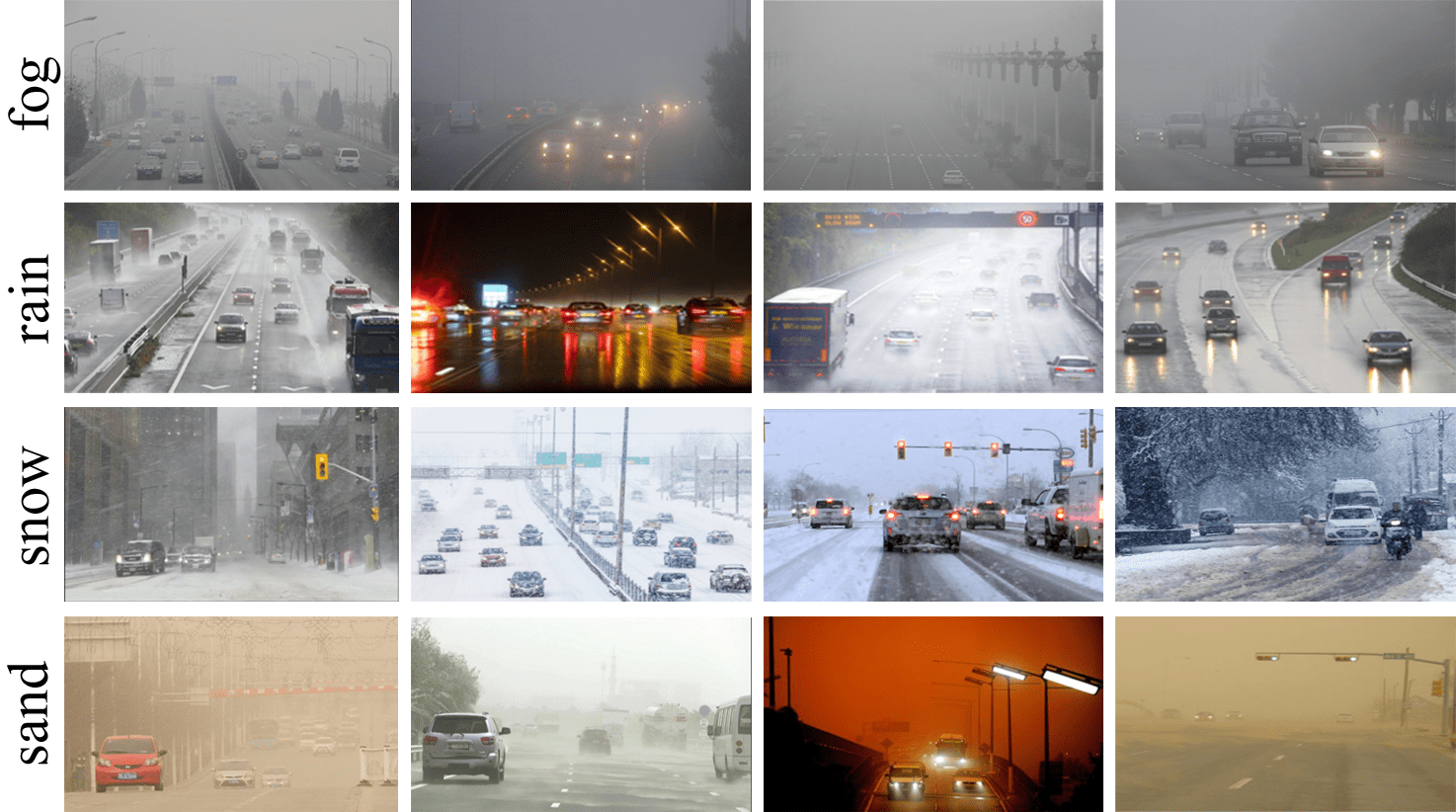}
	\caption{Sample images of the DAWN dataset illustrating four cases of adverse weather conditions.}
	\label{DAWN_ex} 
\end{figure*}
There is a need for a dataset of real-world images addressing the shortcomings of the aforementioned datasets considering imaging in bad weather conditions. Currently, it is uncertain how deep learning algorithms would carry out on the wild through the influence of cross-generalization for adverse weather conditions. In addition, how the progress of these algorithms is standardized and applied safely in the ITS's applications. To this end, we introduce a novel dataset of real-world images collected under various adverse weather conditions, which we called "DAWN: Detection in Adverse Weather Nature". It is designed to support the research in ITS's applications for safety opportunities. The unique characteristics of DAWN dataset give researchers a chance to examine aspects of vehicles detection that have not been examined before in the literature as well as issues that are of key importance for autonomous vehicles technology and ITS safety applications.

\begin{figure*}[!]
\centering
		\includegraphics[width=18 cm,height=!]{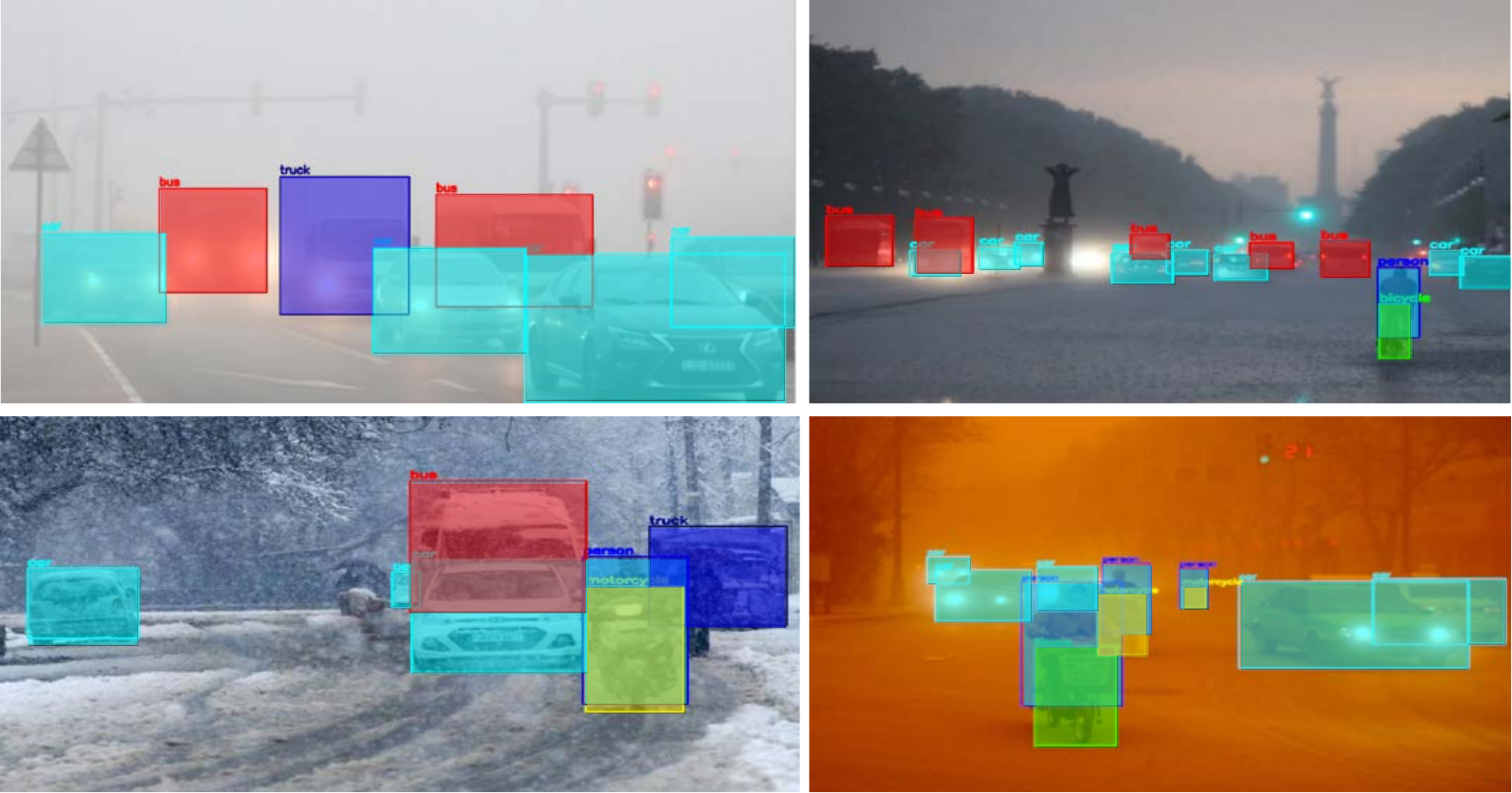}
		\caption{Examples of annotations in DAWN dataset.
		The dataset is annotated using LabelMe \cite{russell2008labelme} into 7,845 total bounding boxes of five types (e.g., car, bus, truck, motorcycles, and bicycles) and person for cyclist/pedestrian.
	}
	\label{DAWN_ann} 
\end{figure*}

The goal of DAWN dataset is to investigate the performance of vehicle detection and classification methods on a wide range of natural images for traffic scenes in the cross-generalization of adverse weather conditions, which are divided into four categories according to the weather (i.e., fog, snow, rain and sand). DAWN dataset contains significant variation in terms of vehicle category, size, orientation, pose, illumination, position and occlusion. Moreover, this dataset exhibits a systematic bias for traffic scenes during nasty winter weather, heavy snow hits, sleet rain, hazardous weather, sand and dust storms. Samples images from DAWN dataset are shown in Figure. \ref{DAWN_ex}. To ensure an accurate evaluation, the traffic scenes are comprehensive with normally moving and congested traffic, combined motorway, highway, urban roads and intersections which built up of several countries to cover the weather change of the different regions in the universe. 
Annotations of the vehicles are consistent, accurate and exhaustive for vehicles' classes (e.g., car, bus, truck, motorcycle and bicycle) with the presence of the human as cyclist and pedestrian. Examples of annotations in DAWN dataset are illustrated in Figure. \ref{DAWN_ann}. 

Images in DAWN dataset are collected through Google and Bing search engines during a visual search that contains a list of query keywords (include; foggy, haze, mist, nasty winter weather, blustery weather, heavy snow hits, sleet rain, sandstorm, duststorm, hazardous weather, adverse weather, traffic, motorway, vehicle). Then, the candidate images are filtered and selected by human in loop. The candidate images for each situation in DAWN must respect the corresponding terms of use for Google, Bing and Flickr terms of use where the license types include: 'Free to share and use'. 
\begin{figure*}[!]
\centering
		\begin{subfigure}{8 cm}\centering
		\includegraphics[width=8 cm,height=!]{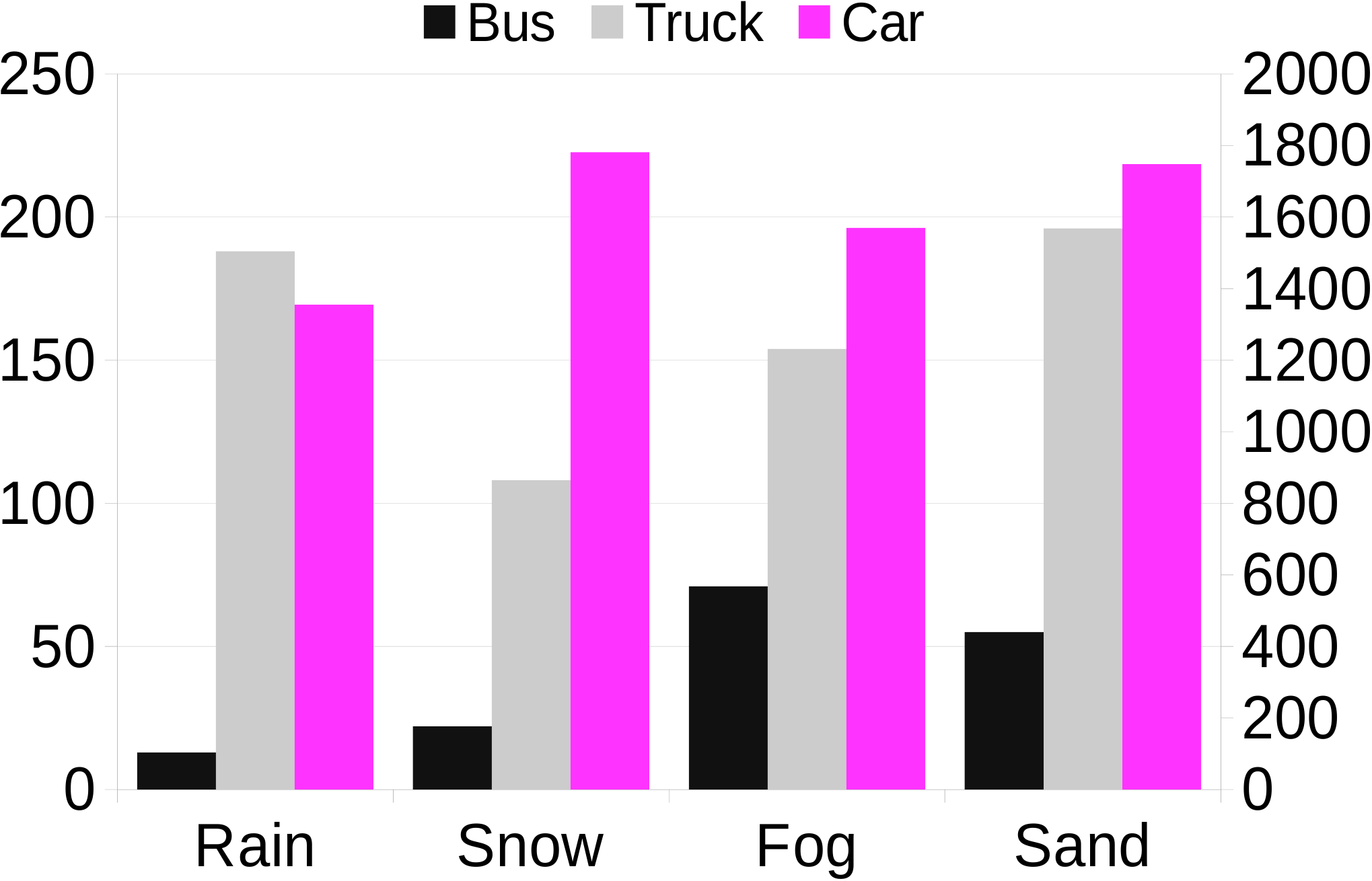}
		\caption{vehicles}
		\end{subfigure} 
		\begin{subfigure}{7.5 cm}\centering
		\includegraphics[width=7.5 cm,height=!]{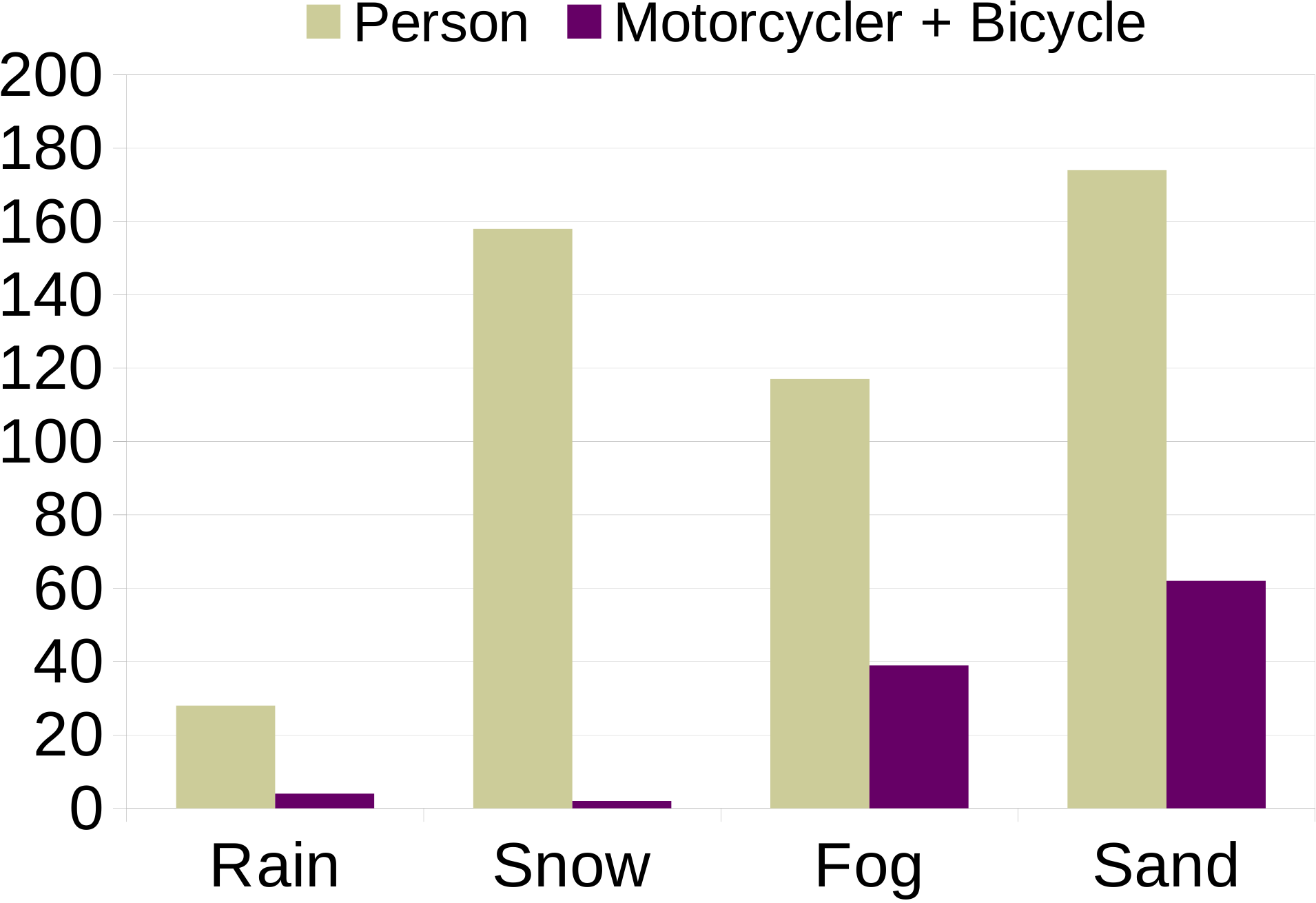}
		\caption{person, motorcycle and bicycle}
		\end{subfigure} 
		\caption{Statistics of DAWN dataset Ground Truth (GT).}
	\label{DAWN_statis} 
\end{figure*}
This dataset contains a collection of 1K image from real-traffic environments, which are divided into four primary subsets according to weather situations: Fog, Rain, Snow and Sand. 
Finally, this dataset is annotated using LabelMe \cite{russell2008labelme} to five types of vehicles and person for cyclist/pedestrian with 7,845 total bounding box of GT including car (82.21\%), bus (2.05\%), truck (8.22\%), motorcycles + bicycles (1.36\%), and person (6.07\%) as reported by charts shown in Figure. \ref{DAWN_statis}.

\section{Conclusion} \label{ConclSec}
In this paper, we proposed a novel dataset (called DAWN) for vehicle detection in adverse weather conditions, including heavy fog, rain, snow and sandstorms. The unique characteristics of the new dataset, DAWN, gives researchers a chance to examine aspects of vehicles detection that have not been examined before in the literature, as well as issues that are of key importance for autonomous vehicles technology and ITS safety applications. 

\section*{Acknowledgment}
The authors would like to thank the providers of vehicle datasets.

\bibliographystyle{IEEEbib}
\bibliography{references}
\vskip -2\baselineskip plus -1fil
\begin{IEEEbiography}[{\includegraphics[width=1in,height=1.25in,keepaspectratio]{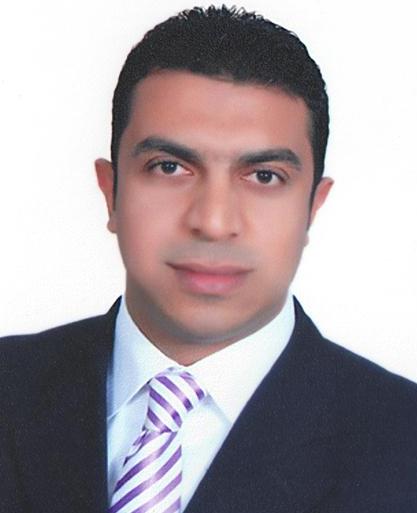}}]{Mourad A. Kenk} received his MSc degree in Computer Science in 2015 from the Faculty of Science, South Valley University. Egypt. He is an assistant lecturer at the Department of Mathematics, Faculty of Science, South Valley University, Egypt. He is currently working toward his Ph.D. degree in Computer Science at the Faculty of Science, South Valley University, Egypt. His research interests include computer vision and robotics. Especially, Object Detection/Tracking, Pose Estimation, Human-aware and Visual Navigation for Autonomous Robots. In 2017, he has joint Electrotechnics and Automatics Research Group (GREAH), Normandy University, Le Havre, France as a researcher. For two years, he worked on Logistics Robotics to develop a picking application in logistics warehouse and intelligent visual surveillance system for mobile robots. He has served as a reviewer for Pattern Recognition Letters and Machine Vision Applications Journals.
\end{IEEEbiography}
\vskip -2\baselineskip plus -1fil
\begin{IEEEbiography}[{\includegraphics[width=1in,height=1.25in,clip,keepaspectratio]{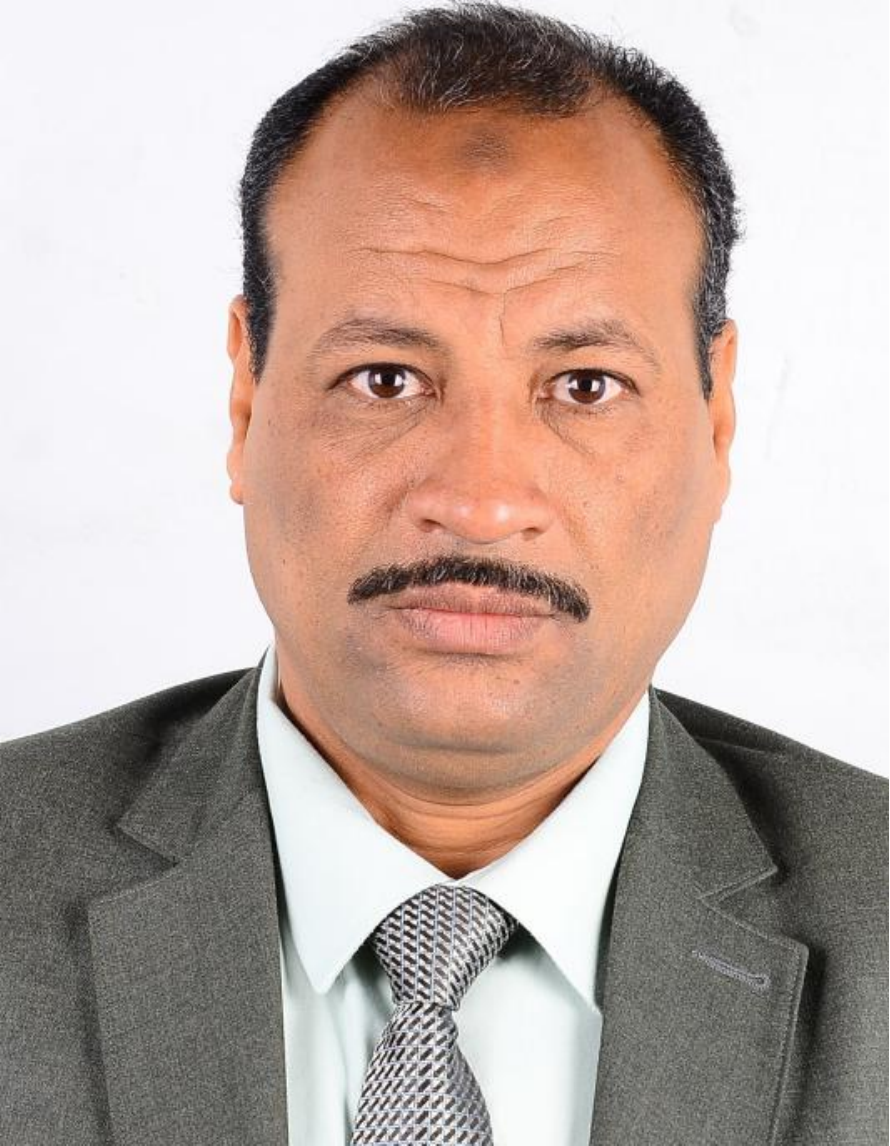}}]{Mahmoud Hassaballah} received his BSc degree in mathematics in 1997 and his MSc degree in computer science in 2003, both from South Valley University, Egypt, and his Doctor of Engineering (DEng) in computer science from Ehime University, Japan in 2011. He is currently an associate professor of computer science at the Faculty of Computers and Information, South Valley University, Egypt. He serves as a reviewer for several journals such as IEEE Transactions on Image Processing, IEEE Transactions on Fuzzy Systems, Pattern Recognition, Pattern Recognition Letters, IET Image Processing, IET Computer Vision, IET Biometrics, Journal of Real-Time Image Processing, and Journal of Electronic Imaging. He has published 5 books and over 50 research papers in refereed journals and conferences. His research interests include feature extraction, object detection/recognition, artificial intelligence, biometrics, image processing, computer vision, machine learning, and data hiding.	
\end{IEEEbiography}
\end{document}